\documentclass[runningheads]{llncs}
\usepackage[T1]{fontenc}
\usepackage{amsmath,amssymb,amsfonts}
\usepackage{graphicx}
\usepackage{subcaption}
\usepackage[absolute]{textpos}

\begin{document}
\begin{textblock}{12}(2,0.3)
	\noindent Please cite as follows: Papatheodoulou, D., Pavlou, P., Vrachimis, S.G., Malialis, K., Eliades, D.G., Theocharides, T. (2022). A Multi-label Time Series Classification Approach for 
	Non-intrusive Water End-Use Monitoring. In: Maglogiannis, I., Iliadis, L., Macintyre, J., Cortez, P. (eds) Artificial Intelligence Applications and Innovations. AIAI 2022. IFIP Advances in Information and Communication Technology, vol 647. Springer, Cham. doi: 10.1007/978-3-031-08337-2\_5
\end{textblock}

\title{A Multi-label Time Series Classification Approach for Non-intrusive Water End-Use Monitoring
\thanks{This work has been supported by the European Union Horizon 2020 program under Grant Agreement No. 739551 (TEAMING KIOS CoE) and the Government of the Republic of Cyprus through the Deputy Ministry of Research, Innovation and Digital Policy, and the FLOBIT project co-funded by the Research and Innovation Foundation of Cyprus, the European Regional Development Fund and Structural Funds of the European Union in Cyprus.}
}
%
%
\author{Dimitris Papatheodoulou\inst{1}\orcidID{0000-0002-1922-1781} \and
Pavlos Pavlou\inst{1}\orcidID{0000-0001-8740-4187} \and
Stelios G. Vrachimis\inst{1,2}\orcidID{0000-0001-8862-5205} \and
Kleanthis Malialis\inst{1,2}\orcidID{0000-0003-3432-7434} \and
Demetrios G. Eliades\inst{1}\orcidID{0000-0001-6184-6366} \and
Theocharis Theocharides\inst{1,2}\orcidID{0000-0001-7222-9152}
}
\authorrunning{D. Papatheodoulou, et al.}
%
\institute{
KIOS Research and Innovation Center of Excellence\\
University of Cyprus, Nicosia, Cyprus\and
Department of Electrical and Computer Engineering\\
University of Cyprus, Nicosia, Cyprus\\
\email{
\{papatheodoulou.dimitris, pavlou.v.pavlos, vrachimis.stelios, malialis.kleanthis, eldemet, ttheocharides\}@ucy.ac.cy
}
}
\maketitle              
\begin{abstract}
Numerous real-world problems from a diverse set of application areas exist that exhibit temporal dependencies. We focus on a specific type of time series classification which we refer to as aggregated time series classification. We consider an aggregated sequence of a multi-variate time series, and propose a methodology to make predictions based solely on the aggregated information. As a case study, we apply our methodology to the challenging problem of household water end-use dissagregation when using non-intrusive water monitoring. Our methodology does not require a-priori identification of events, and to our knowledge, it is considered for the first time. We conduct an extensive experimental study using a residential water-use simulator, involving different machine learning classifiers, multi-label classification methods, and successfully demonstrate the effectiveness of our methodology.

\end{abstract}
\section{Introduction}
The ever-increasing volume of data that is accumulated in various application areas in recent years has accelerated the development of time series methods for forecasting or classification purposes. Examples include critical infrastructure systems, such as, water distribution networks and power load balancing, as well as areas such as healthcare, finance, environmental monitoring, and retail. What real-world problems from the aforementioned areas have in common is the existence of temporal dependencies. One unique attribute of time series data underlies in the chronological order of the observations which constitutes a challenging factor in their analysis. Thus, time series modelling has significant importance as it needs to account for trends, seasonality and abrupt changes that are often exhibited in time series data \cite{fawaz2019deep}.

\par Time series classification is a general task which is useful in numerous areas and applications. It is a type of a supervised machine learning problem, where time series data are described by a class label. The difference with other classification problems is that the natural temporal order in the data is significant, and a learning algorithm has to identify and exploit the temporal characteristics.

In this work, we consider the problem of non-intrusive water usage monitoring in households as our case study \cite{Mazzoni2021}.
As water scarcity is increasingly affecting the world, the development of new strategies focusing on water conservation has become crucial. 
To this direction, many investments have been made the last years in data and information technologies to facilitate the use of smart water meters for domestic use.  
Smart metering of domestic water consumption to continuously monitor the usage of different water fixtures and appliances has been shown to have an impact on people’s behavior towards water conservation \cite{cominola2021long}. 
However, the installation of multiple sensors to monitor each appliance may have a high initial cost and there is currently no simple and cost-effective method to monitor end-use water consumption.
This work aims to address this issue by identifying active water consuming fixtures and appliances using only data from the main water flow meter, measuring the total household consumption.

Towards this direction, studies have focused on using measurements of the total domestic consumption with Machine Learning (ML) methods to disaggregate water usage into each appliance \cite{Nguyen2013}. 
Identifying which appliances are in use through ML is challenging since their operation may be overlapping, while specific appliances may operate with intermittent flow, making individual consumption events hard to distinguish. 
Decision trees and Machine learning algorithms with a combination of unsupervised learning with feature extraction and clustering evaluation have been identified as the major disaggregation techniques to classify water end-use appliances using total consumption data \cite{cominola2015benefits}. 
The authors in \cite{froehlich2011longitudinal} used a Bayesian approach coupled with a template classifier, a language model, grammar and prior probabilities to classify water events, using however additional measurements from pressure sensors.
The study achieved 90\% and 94\% accuracy at fixture level based on data collected form five sites during a five-week period. 
An adaptable neuro fuzzy network called Anfis has been proposed in \cite{ojeda2008classification} using two approaches to estimate each class: the space and the mean values and standard deviation of each class. 
This method achieved a 91\% score on classifying water end-uses using a limited dataset of flow measurement from one point in a single house.
The authors in \cite{Mazzoni2021} propose a rule-based methodology for end-use disaggregation, by prioritizing the identification of appliances with comparably regular behavior.
The methodology was applied to a sample of four households in Italy where detailed water-use data were collected at the inlet point and at each end-use over a period of 2 months.
The results were consistent with those obtained in similar studies making use of synthetic data \cite{cominola2018implications}.
In \cite{Charalampous2021}, the authors examine the use of a novel flow indicator to deal with simultaneous water consumption events.
Six water usage features are extracted from each event : duration, volume, flow peak, mode, time of day, and day of week. 
A Random Forest classifier is used to identify the category of each event based on the extracted features. 
It is a common characteristic of the aforementioned studies to use pre-processing algorithms to identify consumption events before classification. This work makes two key contributions as follows:
\begin{enumerate}
    \item We consider a specific type of time series classification which we refer to as aggregated time series classification. This setting is largely unexplored in the literature, and considers an aggregated sequence of a multi-variate time series. We propose a methodology to make predictions based solely on the aggregated information.
    
    \item As a case study we consider the challenging problem of non-intrusive water end-use monitoring, where the proposed methodology, which uses only a sliding window of measurements and does not require \textit{a-priori} identification of events, is considered for the first time, to the authors knowledge.
\end{enumerate}

The paper is organised as follows. We formulate the problem in Section~\ref{sec:formulation}. The experimental setup is described in Section~\ref{sec:setup}. An empirical analysis of the proposed methodology is provided in Section~\ref{sec:empirical_analysis}, while in Section~\ref{sec:comparative_study} we compare the performance of various learning algorithms. We conclude in Section~\ref{sec:conclusions}.

\section{Problem Formulation}\label{sec:formulation}

We consider a data generating process that provides at each time step $t$ a sequence of instances $S = \{(x^t,y^t)\}_{t=1}^T$ from an unknown probability distribution $p^{t}(x,y)$, where $T \in [1,\infty)$.

The input $x^t = \{x^t_i\}_{i=1}^d \in \mathbb{R}^d$ is a $d$-dimensional vector belonging to input space $X \subset \mathbb{R}^d$. The instances constitute a \textbf{multivariate} time series with $d$ number of time series, and each corresponds to a \textbf{univariate} time series defined as $z_i = \{x^t_i\}_{t=1}^T \in \mathbb{R}^T$.

The label (i.e., the ground truth) of the classification task is denoted by $y^t \in Y$. When $Y = \{1, ..., K\}$, $K \geq 2$, it is termed \textbf{multi-class} classification, that is, it refers to a task with more than two classes. When $Y \in \{0,1\}^K$ it is termed \textbf{multi-label} classification, i.e., it assigns to each instance a set of labels. Each digit corresponds to the inclusion (1) or absence (0) of the relevant label.

A classifier $h$ receives a new example $x^t$ at time step $t$ and makes a prediction $\hat{y}^t$ based on a concept $h: X \to Y$ such that $\hat{y}^t = h(x^t)$. We refer to this as \textbf{time series classification}. To capture the temporal aspects of the data, it is possible to introduce a memory component, such as, a sliding window to facilitate with the prediction task, i.e., $\hat{y}^t = h(x^t, x^{t-1}, ..., x^{t-W})$, where $W$ is the window size.

Let us define the aggregated sequence ${x^*}^t = \{x^t_1 \otimes x^t_2 \otimes ... \otimes x^t_d$\}, where the classifier makes a prediction based solely on the aggregated sequence ${x^*}^t$ at time step $t$, where $\otimes$ is an aggregation operator depending on the application. We refer to this as \textbf{aggregated time series classification}.


We formulate the non-intrusive water end-use monitoring problem as an aggregated time series classification task. For the generation of the dataset, we use a residential water demand simulator \cite{cominola2018implications} (descibed below) that synthesises a sequence of instances $S$ at each time step $t$ based on the water consumption profiles of appliances in U.S. households.

The generated time series consists of five signals, that correspond to the water flow consumption of the toilet ($z_1$), shower ($z_2$), faucet ($z_3$), clothes washer ($z_4$) and dish washer ($z_5$). From the consumption of these five appliances, we compose an aggregated sequence  ${x^*}^t$, by summing the water flow consumption from each sequence $z_i$ at time step $t$. The aggregated sequence ${x^*}^t$ is described by a set of binary labels that correspond to the appliances that were active (1) or inactive (0) at each time step $t$. To improve the efficacy of the predictions, we use a sliding window approach that allows the classifier to capture information from previous time steps, thus extracting underlying temporal patterns.

\section{Experimental Setup}\label{sec:setup}

\begin{figure}[t!]
    \centering
    \includegraphics[width=0.5\linewidth]{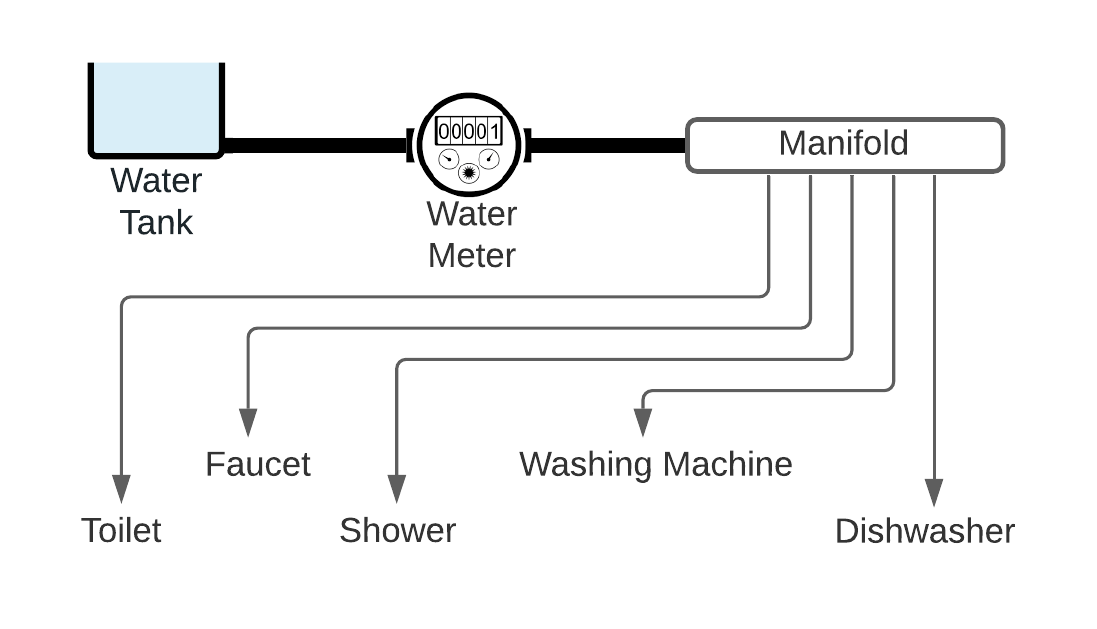}
    \caption{Piping connectivity in a domestic water system.}%
    \label{fig:pipingConfiguration}%
\end{figure}

\subsection{Simulator and dataset}
This case study uses the STochastic Residential water End-use Model (STREaM) \cite{cominola2018implications}, a modelling software developed to generate synthetic time series data of a household with resolution of up to 10s. STREaM generates time series of each water end-use fixture characterised by its signature (i.e., typical consumption pattern), as well as its probability distributions of number of uses per day, single use durations, water demand contribution, and time of use during the day. STREaM takes into consideration the number of house occupants in the calculation of total household water demand. STREaM was calibrated on a large dataset including observed and disaggregated water end-uses from over 300 single-family households in nine U.S. cities \cite{deoreo2011analysis}. The following water end-uses were considered in this dataset: toilet, shower, faucet, clothes washer, dishwasher. The end-uses are further distinguished in standard and high efficiency appliances, which have different consumption characteristics. Each fixture can be activated only once during each time step but multiple fixtures can be active during the same time. The dataset provides the water flow reading at each time step for each fixture and their sum as the total consumption. Figure \ref{fig:pipingConfiguration}, illustrates the connectivity of a domestic water system. The main flow meter is located at the outflow of a domestic house water tank thus allowing us to measure water flow for each time step. The system consists of the main water tank, the main water pipe connecting the tank to the manifold, the manifold which distributes the water to the house piping system and the end-use appliances.

\begin{figure}[t!]
	\centering
	
	\subfloat[Toilet]{\includegraphics[scale=0.26]{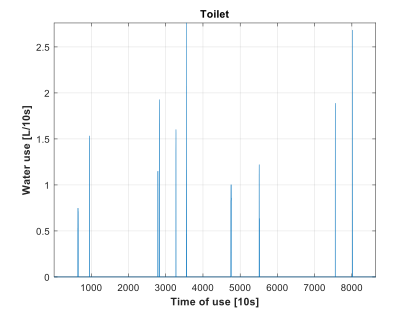}%
		\label{fig:dataset_toilet}}
	\subfloat[Shower]{\includegraphics[scale=0.26]{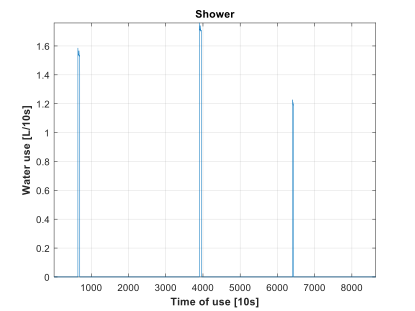}%
		\label{fig:dataset_shower}}
	\subfloat[Faucet]{\includegraphics[scale=0.26]{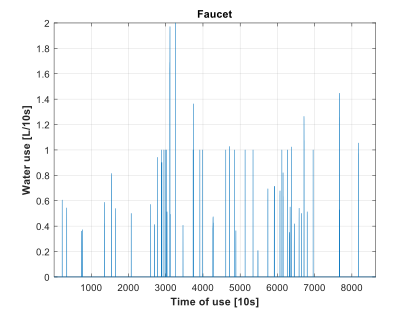}%
		\label{fig:dataset_faucet}}
		
	\subfloat[Washer]{\includegraphics[scale=0.26]{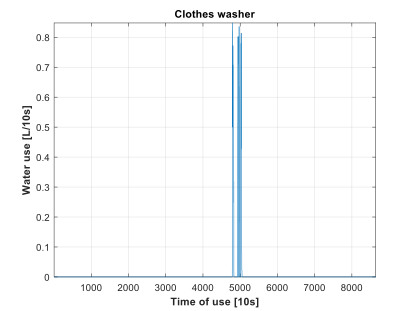}%
		\label{fig:dataset_washer}}
	\subfloat[Dishwasher]{\includegraphics[scale=0.26]{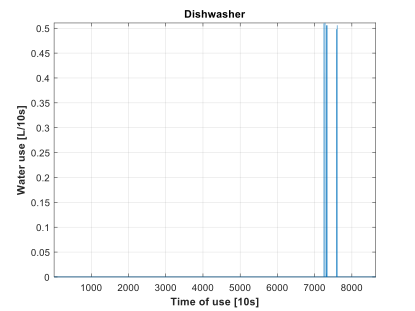}%
		\label{fig:dataset_dishwasher}}
	\subfloat[Total]{\includegraphics[scale=0.26]{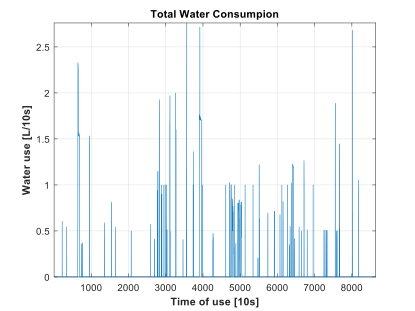}%
		\label{fig:dataset_total}}
	
	\caption{Household water consumption per appliance}
\end{figure}

\begin{figure}[h!]
    \centering
    \subfloat[\centering None instances]{{\includegraphics[scale=0.16]{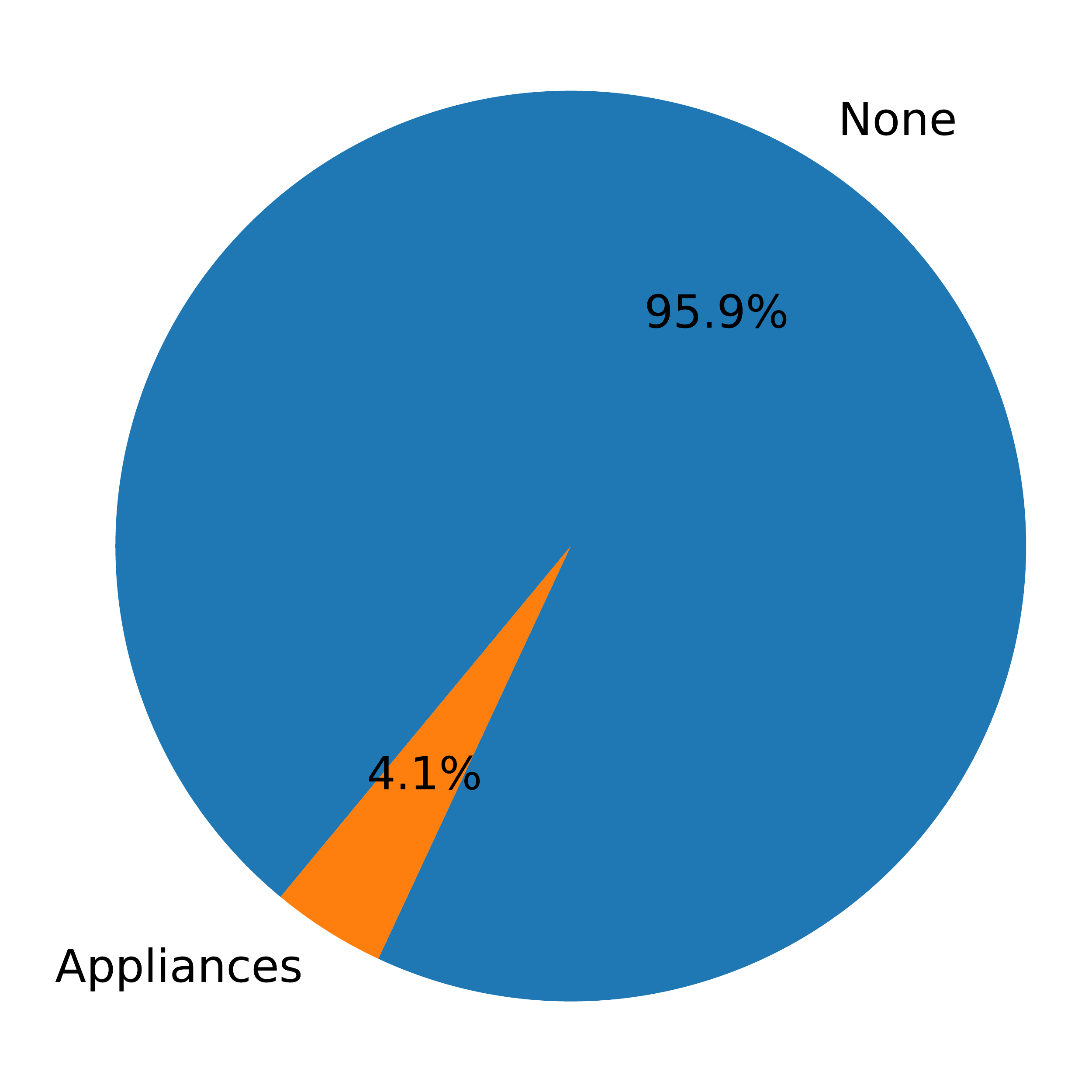} }}%
    \subfloat[\centering Appliance instances]{{\includegraphics[scale=0.20]{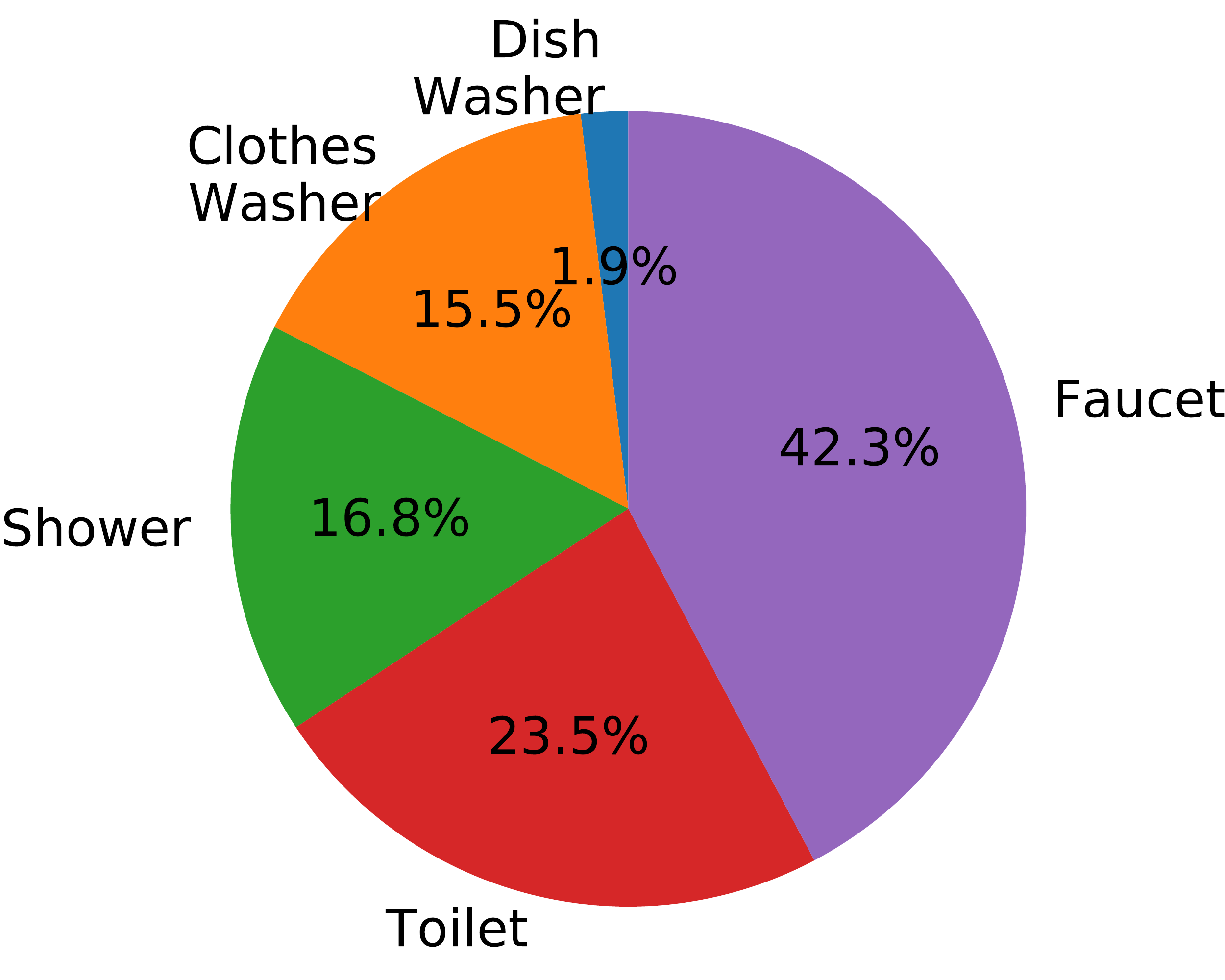} }}%
    
    \caption{Dataset's class sizes}%
    \label{fig:dataset_appliance_counts}%
\end{figure}

The dataset used in this study considers the use of standard toilet, standard shower, standard faucet, high efficiency clothes washer and standard dishwasher in a 2-person household for a period of 180 days (6 months) and it has a resolution of 10s. Figures~\ref{fig:dataset_toilet}-\ref{fig:dataset_dishwasher} depict the household water consumption for each appliance for a random day. Figure~\ref{fig:dataset_total} shows the total water consumption. Figure~\ref{fig:dataset_appliance_counts} displays each class' size (i.e., time series records), where it is clear that imbalance is severe.

We split the dataset into three subsets, 3-months worth of data are held for training, and two sets of 1.5 months of historical data are reserved for validation and test sets. That equates to a set of 777600 samples for the training and 388800 samples for both validation and test sets. The training subset consists of the samples that are given to the model, to identify and learn any underlying patterns of the data. The validation subset contains data that is used for evaluation purposes in order to optimize the model. Lastly, the test subset is a set of unseen samples that are used only to assess the performance of the algorithms, to determine how well the algorithms can generalize on unseen data.

\subsection{Classification algorithms}

\textbf{Random Forest (RF)} \cite{bishop2006}:
It is a tree-based, ensemble learning algorithm, i.e., it depends on multiple tree-based learners which make individual predictions that are then averaged together. Typically, the more trees it has, the more robust model it is as its performance does not rely on a single tree.

\textbf{Extreme Gradient Boosting (XGBoost)} \cite{chen2016xgboost}:
It is a machine learning technique that produces a prediction model in the form of an ensemble of weak prediction models, which are typically tree-based. This technique builds a model in a stage-wise fashion and combines weak learners into a single strong learner. As each weak learner is added, a new model is fitted to provide a more accurate estimation. The XGBoost classifier is a tree-based ensemble machine learning algorithm with Gradient Boosting as its main component. Moreover, XGBoost has the ability to handle missing values on its own and it is very effective and efficient in terms of performance as well as training time even on large datasets.

\textbf{Multilayer Perceptron (MLP)} \cite{bishop2006}:
It is a feed-forward neural network that consists of an input and an output layer, and can have multiple hidden layers. MLP uses the backpropagation algorithm for training which computes the gradient of the loss function with respect to the weights of the neural network.

\subsection{Multi-task classification methods}
Multi-task classification \cite{caruana1997multitask} combines (related or unrelated) tasks by using the principles of transfer learning. It focuses on the preservation of the knowledge gained while solving a particular task and then applying it to a different task. It improves the generalisation by associating information of multiple tasks. The learning process happens simultaneously across all tasks while using shared representation, which may lead to performance improvements. In this work we use the terms multi-task and multi-label classification interchangeably.

Meta-estimators have the flexibility to appropriately leverage all the characteristics of a base estimator in multiple forms. They can utilize the information of each individual base estimator in a way that extends their capabilities. The following meta-estimators describe how they can consolidate a multi-class binary classification problem into a more generalised version, a multi-task problem.

\begin{itemize}
    \item \textbf{Binary Relevance (BR)} \cite{luaces2012binary}: It is a problem transformation technique, where each label is treated separately, as a binary classification problem. Thus, the multi-task problem is split into binary classification sub-tasks, where any type of supervised classification algorithm can be defined as the base-estimator. Subsequently, the meta-estimator is constructed that fits each individual base-estimator in order to optimize its loss function. The predictions are then combined into a multi-output format.
    
    \item \textbf{Classifier Chain (CC)} \cite{read2011classifier}: It operates similarly to Binary Relevance, however, it is capable of exploiting correlations among target variables. The difference between this approach and BR is that in a multi-label classification setting with $N$-classes, $N$-binary classifiers are assigned a number that corresponds to their order in the classifier’s chain. The training process follows the order of the models in the chain, where each binary classifier is fit on the available training data with the addition of the actual target labels of the classes whose models were assigned a lower order in the chain.
\end{itemize}

\subsection{Evaluation metrics}
Classifiers are typically evaluated using the \textbf{accuracy} metric.
However, this metric becomes unsuitable as it is biased towards the majority (normal) class. A widely accepted metric which is less sensitive to imbalance is \textbf{F1-Score}, defined below as the harmonic mean of the model's precision and recall \cite{he2009learning}. For multi-label classification, we will use the micro-averaging F1-score or \textbf{F1-Micro}.

\begin{equation}\label{eq:f1}
    F1 = 2 \times \frac{precision \times recall}{precision + recall} 
\end{equation}




\section{Empirical Analysis}\label{sec:empirical_analysis}

\subsection{Hyper-parameter tuning}
We have tuned all the algorithms (RF, XGBoost, MLP) using both multi-task methods (BR, CC). We have also used four different sliding windows that capture the previous 60, 120, 240 and 480 time steps that correspond to 10, 20, 40 and 80-minute intervals respectively. To facilitate the reproducibility of our results, this section provides the values of the hyper-parameters after tuning. Due to space restrictions, we present the hyper-parameter values using the CC multi-label method. Table \ref{tab:random_forest_cc_tuned_params} shows the hyper-parameters which yield the best performance for the Random Forest using the CC multi-task method. Table \ref{tab:xgboost_cc_tuned_params} shows the hyper-parameters which yield the best performance for XGBoost using the CC multi-task method. Table \ref{tab:mlp_cc_tuned_params} shows the hyper-parameters which yield the best performance for MLP using the CC multi-task method.


\begin{table}[h]
\centering
\resizebox{\columnwidth}{!}{%
\begin{tabular}{lcccc}
\hline
\textbf{} & \textbf{Window 60} & \textbf{Window 120} & \textbf{Window 240} & \textbf{Window 480} \\ \hline
\textbf{Number of estimators} & 325 & 225 & 475 & 375 \\
\textbf{Criterion} & Gini & Gini & Gini & Entropy \\
\textbf{Max Depth} & 8 & 8 & 9 & 9 \\
\textbf{Max Features} & Auto & Auto & Sqrt & Sqrt \\
\textbf{Class Weight} & Balanced & Balanced & Balanced & Balanced \\ \hline
\end{tabular}%
}
\caption{Tuned hyper-parameter values for RF (CC)}
\label{tab:random_forest_cc_tuned_params}
\end{table}
 

\begin{table}[h]
\centering
\resizebox{\columnwidth}{!}{%
\begin{tabular}{lcccc}
\hline
\textbf{} & \textbf{Window 60} & \textbf{Window 120} & \textbf{Window 240} & \textbf{Window 480} \\ \hline
\textbf{Number of estimators} & 275 & 100 & 275 & 125 \\
\textbf{Max Depth} & 3 & 10 & 3 & 6 \\
\textbf{Learning rate} & 0.05 & 0.03 & 0.05 & 0.03 \\
\textbf{Booster} & Dart & Dart & Gbtree & Gbtree \\
\textbf{Subsample} & 0.4 & 0.1 & 0.7 & 0.2 \\
\textbf{Colsample by tree} & 0.3 & 0.7 & 0.7 & 0.7 \\
\textbf{Colsample by level} & 0.8 & 0.4 & 0.5 & 0.6 \\
\textbf{Colsample by node} & 0.9 & 0.7 & 0.6 & 0.8 \\
\textbf{L1 Regularisation} & 0.03 & 0.06 & 0.02 & 0.03 \\
\textbf{L2 Regularisation} & 0.05 & 0.05 & 0.02 & 0.0006 \\ \hline
\end{tabular}%
}
\caption{Tuned hyper-parameters values for XGBoost (CC)}
\label{tab:xgboost_cc_tuned_params}
\end{table}


\begin{table}[h]
\centering
\resizebox{\columnwidth}{!}{%
\begin{tabular}{lcccc}
\hline
\textbf{} & \textbf{Window 60} & \textbf{Window 120} & \textbf{Window 240} & \textbf{Window 480} \\ \hline
\textbf{\#hidden layers} & 2 & 2 & 3 & 2 \\
\textbf{\#hidden units} & 32 & 32 & 16 & 64 \\
\textbf{Activation} & Tanh & Tanh & Tanh & Tanh \\
\textbf{Epochs} & 100 & 75 & 175 & 125 \\
\textbf{Optimizer} & Adam & SGD & SGD & SGD \\
\textbf{Learning rate} & 0.01 & 0.05 & 0.09 & 0.04 \\
\textbf{Batch size} & 256 & 256 & 128 & 32 \\
\textbf{L2 Regularisation} & 0.04 & 0.07 & 0.06 & 0.01 \\ \hline
\end{tabular}%
}
\caption{Tuned hyper-parameters values for MLP (CC)}
\label{tab:mlp_cc_tuned_params}
\end{table}

\subsection{Role of the sliding window size}
We now examine the role of the window size. Four different sliding window sizes are examined that capture the previous 60, 120, 240 and 480 time steps that correspond to 10, 20, 40 and 80-minute intervals respectively. In all experiments, we have used the models that yielded the best performance after tuning. Due to space constraints, we present the results only for the CC method.


\textbf{Random Forest}:
Table \ref{tab:random_forest_cc_performance} shows the RF's performance using the CC multi-task method for different window sizes. It can be observed that the performance of the Random Forest declines as the sliding window size becomes larger. The best performance is obtained when the window size is 60. 

\textbf{XGBoost}:
Table \ref{tab:xgboost_cc_performance} shows the XGBoost's performance using CC for different window sizes. Its performance improves as the sliding window grows, for instance, the highest performance in Table~\ref{tab:xgboost_cc_performance} is obtained when the window size is 480. XGBoost can better capture the time correlations in the time-series data.

\textbf{MLP}:
Table \ref{tab:mlp_cc_performance} show the MLP's performance using CC for different window sizes. As with XGBoost, the models' performance increases when the window size becomes larger, however, after some point its performance declines. The best performance is obtained when the size is 120.

Overall, the role of the sliding window is very important as it appears to affect the performance of all models. Overall, a larger window size helps a model capture time correlations in time-series data, however, the performance may start to decline after very large windows as in the case of MLP. Moreover, some models fail to capture time correlations in this domain area despite the increase in the window size; this has been the case with the RF model. While the above serve as guidelines, tuning the sliding window size is necessary.


\begin{table}[h!]
\centering
\resizebox{\columnwidth}{!}{%
\begin{tabular}{lcccc}
\hline
 & \textbf{Window 60} & \textbf{Window 120} & \textbf{Window 240} & \textbf{Window 480} \\ \hline
\textbf{Accuracy} & 96.51 & 96.43 & 96.18 & 96.44 \\
\textbf{F1-Micro} & \textbf{55.75} & 54.22 & 52.39 & 48.60 \\ \hline
\end{tabular}%
}
\caption{Performance of Random Forest (CC)}
\label{tab:random_forest_cc_performance}
\end{table}


\begin{table}[h!]
\centering
\resizebox{\columnwidth}{!}{%
\begin{tabular}{lcccc}
\hline
 & \textbf{Window 60} & \textbf{Window 120} & \textbf{Window 240} & \textbf{Window 480} \\ \hline
\textbf{Accuracy} & 98.80 & 98.86 & 98.78 & 98.76 \\
\textbf{F1-Micro} & 68.91 & 71.94 & 71.98 & \textbf{72.38} \\ \hline
\end{tabular}%
}
\caption{Performance of XGBoost (CC)}
\label{tab:xgboost_cc_performance}
\end{table}


\begin{table}[h!]
\centering
\resizebox{\columnwidth}{!}{%
\begin{tabular}{lcccc}
\hline
 & \textbf{Window 60} & \textbf{Window 120} & \textbf{Window 240} & \textbf{Window 480} \\ \hline
\textbf{Accuracy} & 98.48 & 98.62 & 98.47 & 98.31 \\
\textbf{F1-Micro} & 62.70 & \textbf{65.82} & 64.87 & 61.58 \\ \hline
\end{tabular}%
}
\caption{Performance of MLP (CC)}
\label{tab:mlp_cc_performance}
\end{table}

\subsection{Role of the multi-label classification method}
We examine now the role of the multi-task methods Binary Relevance (BR) and Classifier Chain (CC). We set the window sizes that yielded the best performance earlier, and we compare which method is more effective. For the Random Forest, we set the window size to 60. For the MLP, we set the window size to 120. For the XGBoost, we set the window size to 240, even though one of the models achieved better performance on larger window size (Table~\ref{tab:xgboost_cc_performance}). This decision was based on the negligible difference of the two models, and due to the lower dimensionality of the dataset which results in a more simplified and efficient model.

\textbf{Random Forest}:
Table \ref{tab:random_forest_multi-task_comparison} compares the performance of the Random Forest with a fixed sliding window size of 60, using BR and CC. The results indicate that CC outperforms BR with a considerable performance improvement.

\textbf{XGBoost}:
Table \ref{tab:xgboost_multi-task_comparison} presents the role of the multi-task method using XGBoost on the dataset with a fixed sliding window size of 240. Similarly, with the Random Forest model, the XGBoost using CC achieves a better performance.

\textbf{MLP}:
Table \ref{tab:mlp_multi-task_comparison} shows the comparison between the Binary Relevance and the Classifier Chain methods using the MLP model with a dataset that has a fixed sliding window size of 120. Likewise, with the previous models, the Classifier Chain approach proves to be more effective using the MLP classifier.

\begin{table}[h!]
\centering
\resizebox{\columnwidth}{!}{%
\begin{tabular}{lcc}
\hline
 & \textbf{Random Forest (BR) - Window 60} & \textbf{Random Forest (CC) - Window 60} \\ \hline
\textbf{Accuracy} & 97.04 & 96.51 \\
\textbf{F1-Micro} & 49.92 & \textbf{55.75} \\ \hline
\end{tabular}%
}
\caption{Comparison of multi-task methods with RF}
\label{tab:random_forest_multi-task_comparison}
\end{table}

\begin{table}[h!]
\centering
\resizebox{\columnwidth}{!}{%
\begin{tabular}{lcc}
\hline
 & \textbf{XGBoost (BR) - Window 240} & \textbf{XGBoost (CC) - Window 240} \\ \hline
\textbf{Accuracy} & 98.37 & 98.78 \\
\textbf{F1-Micro} & 70.14 & \textbf{71.98} \\ \hline
\end{tabular}%
}
\caption{Comparison of multi-task methods with XGBoost}
\label{tab:xgboost_multi-task_comparison}
\end{table}

\begin{table}[h!]
\centering
\resizebox{\columnwidth}{!}{%
\begin{tabular}{lcc}
\hline
 & \textbf{MLP (BR) - Window 120} & \textbf{MLP (CC) - Window 120} \\ \hline
\textbf{Accuracy} & 98.29 & 98.62 \\
\textbf{F1-Micro} & 64.21 & \textbf{65.82} \\ \hline
\end{tabular}%
}
\caption{Comparison of multi-task methods with MLP}
\label{tab:mlp_multi-task_comparison}
\end{table}

\begin{table}[h!]
\centering
\resizebox{\columnwidth}{!}{%
\begin{tabular}{lccc}
\hline
 & \textbf{\begin{tabular}[c]{@{}c@{}}Random Forest (CC) –\\ Window 60\end{tabular}} & \textbf{\begin{tabular}[c]{@{}c@{}}XGBoost (CC) –\\  Window 240\end{tabular}} & \textbf{\begin{tabular}[c]{@{}c@{}}MLP (CC) –\\ Window 120\end{tabular}} \\ \hline
\textbf{Accuracy} & 96.51 & 98.78 & 98.62 \\
\textbf{F1-Micro} & 55.75 & \textbf{71.98} & 65.82 \\ \hline
\end{tabular}%
}
\caption{Comparison of different classifiers}
\label{tab:classifier_comparison}
\end{table}

Overall, we can conclude that the Classifier Chain approach proved to be more effective on all occasions that were tested. It had a drastic impact on the improvement of the performance of the Random Forest method, but it also demonstrated a noticeable improvement for both XGBoost and MLP models.

\section{Comparative Study}\label{sec:comparative_study}
We now examine the performance of each classifier. We keep the same window sizes identified previously, and we set the multi-task method to the Classifier Chain as it outperformed the Binary Relevance method on every experimental series. Table \ref{tab:classifier_comparison} compares the performance of each classifier, where it is evident that the XGBoost model significantly outperforms the other classifiers.

\begin{figure}[t!]
    \centering
    
	\subfloat[Toilet]{\includegraphics[scale=0.26]{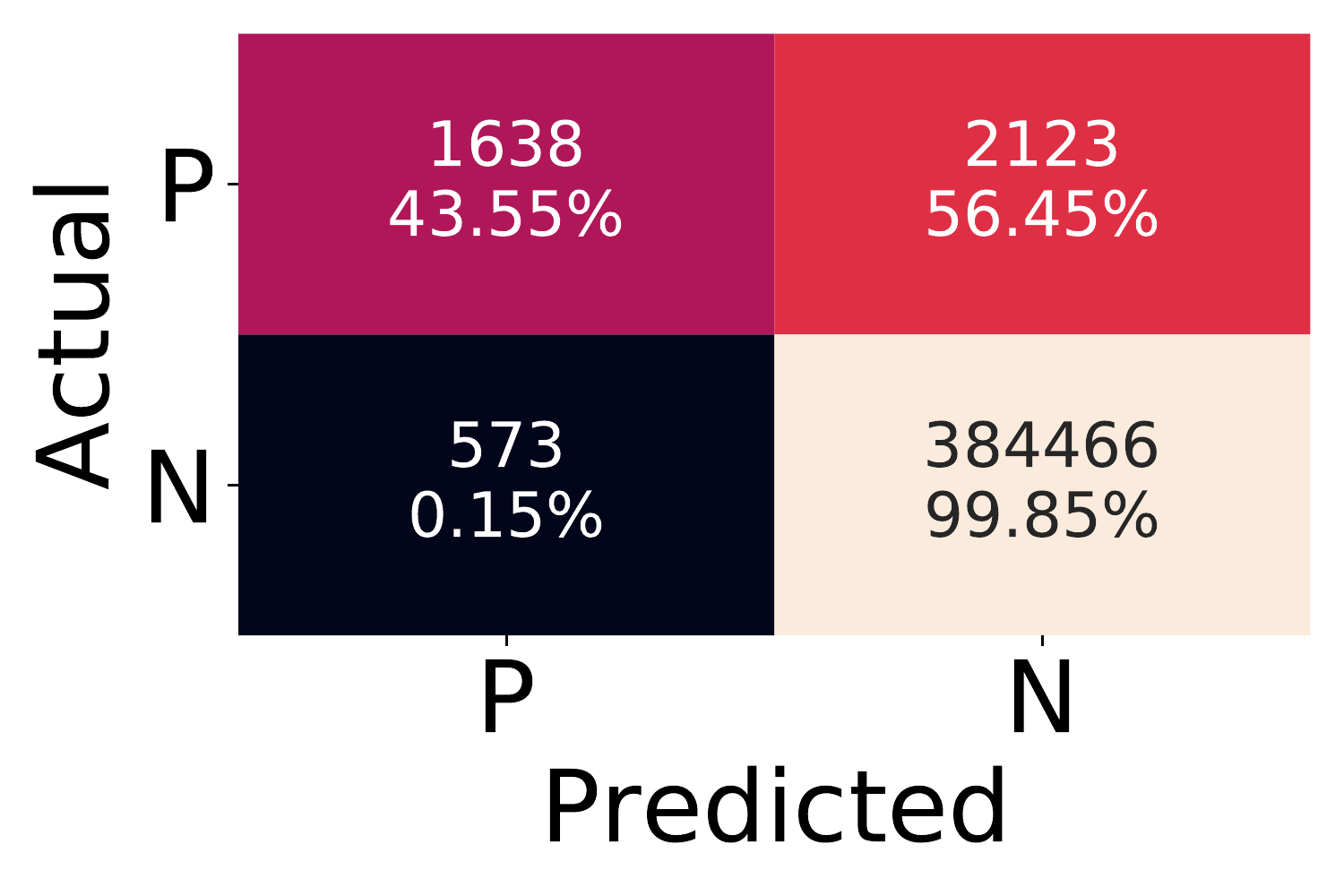}%
		\label{fig:confusion_matrix_toilet}}
	\subfloat[Shower]{\includegraphics[scale=0.26]{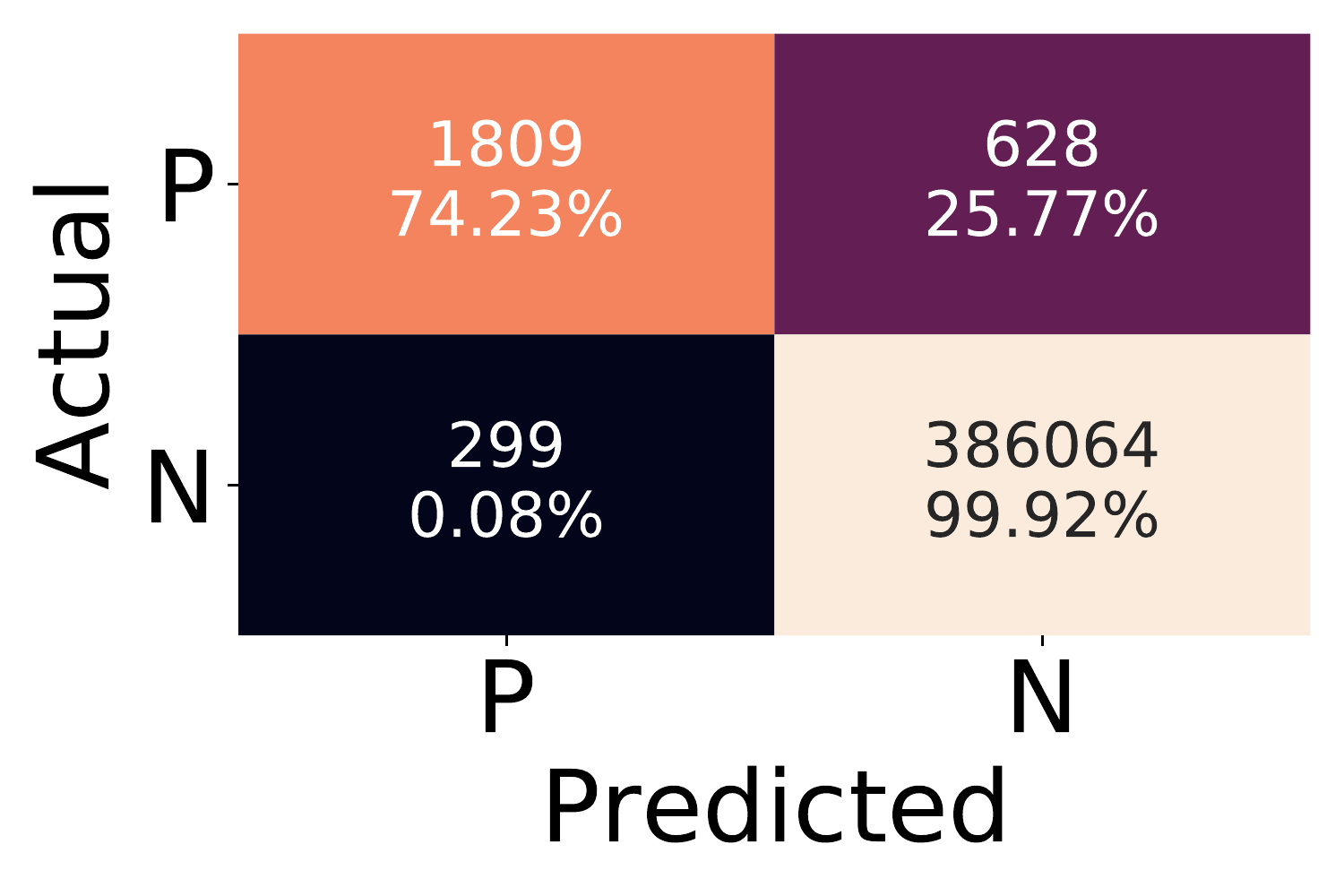}%
		\label{fig:confusion_matrix_shower}}
	\subfloat[Faucet]{\includegraphics[scale=0.26]{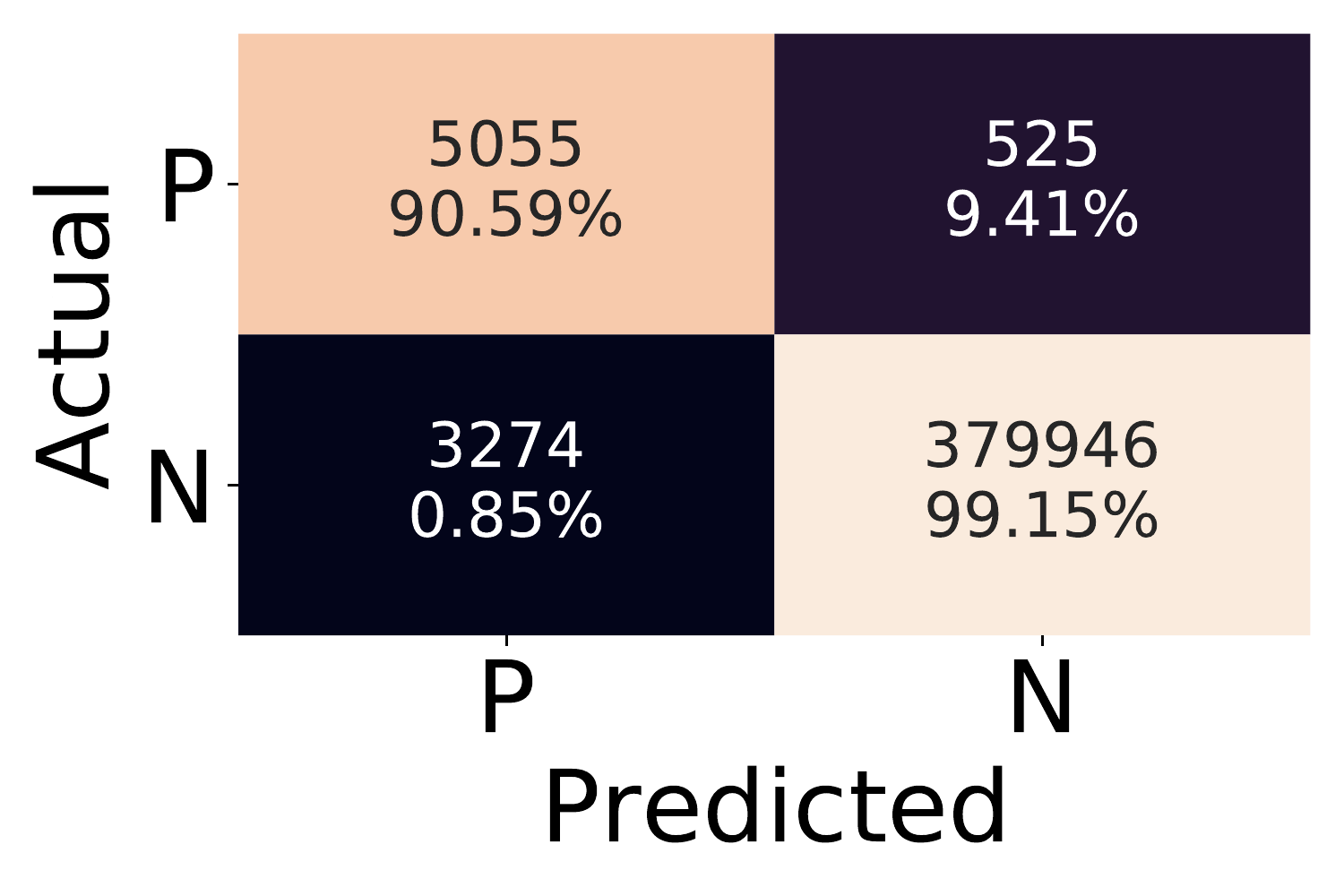}%
		\label{fig:confusion_matrix_faucet}}
		
	\subfloat[Dish washer]{\includegraphics[scale=0.26]{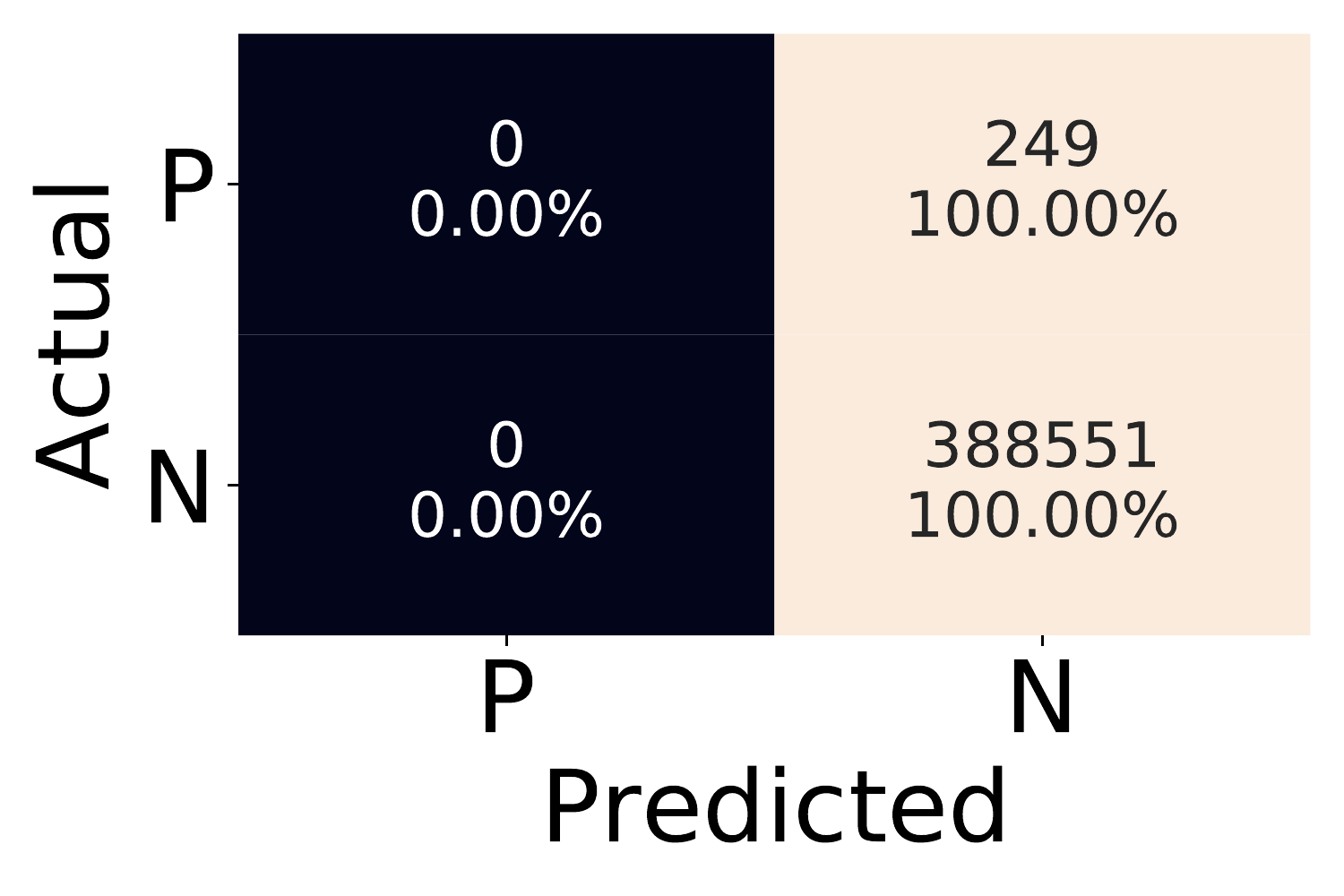}%
		\label{fig:confusion_matrix_dish_washer}}
	\subfloat[Clothes washer]{\includegraphics[scale=0.26]{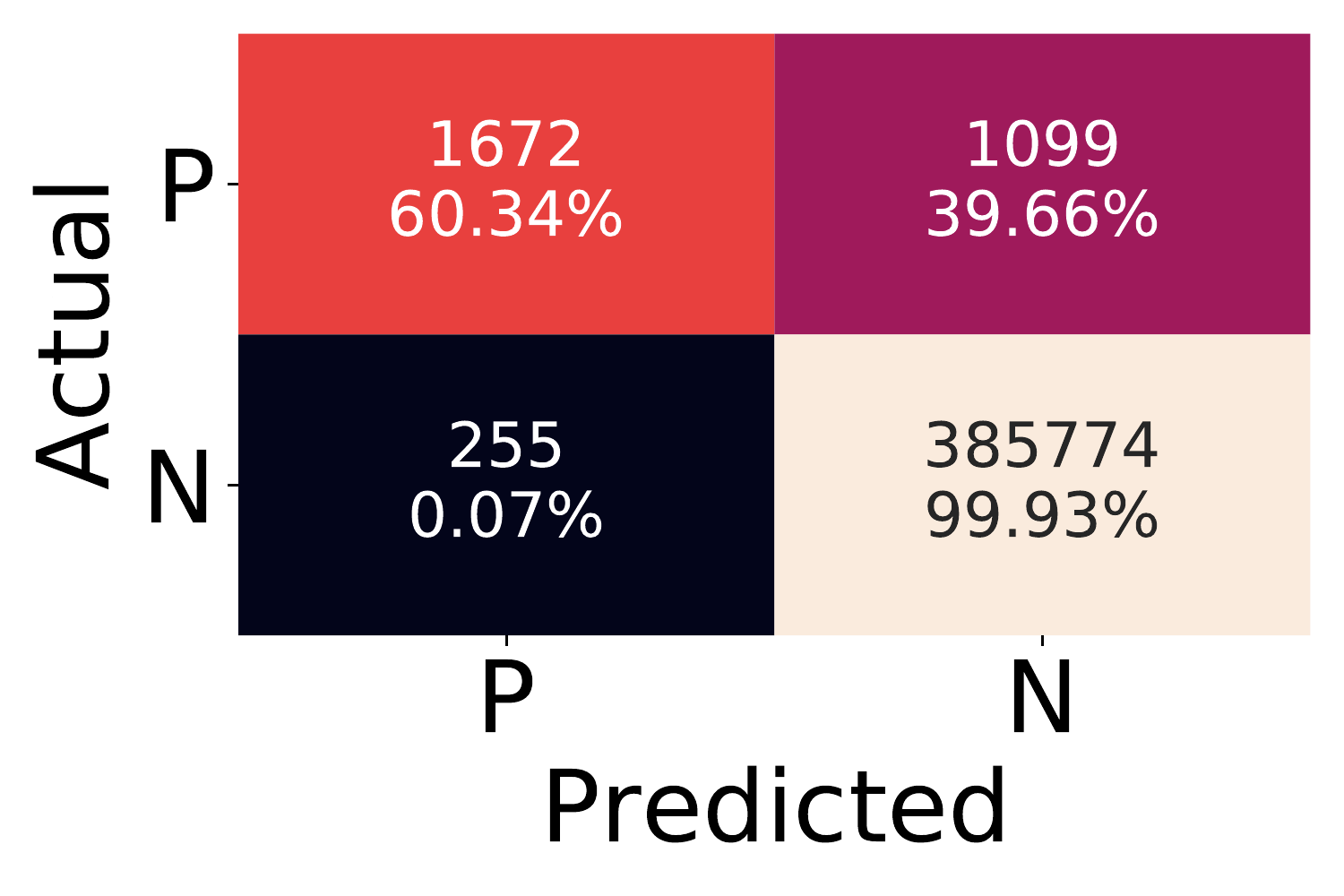}%
		\label{fig:confusion_matrix_clothes_washer}}
	
	\caption{Confusion matrices}%
\end{figure}

Figures \ref{fig:confusion_matrix_toilet} - \ref{fig:confusion_matrix_clothes_washer}, depict the confusion matrices of XGBoost (CC) with a window of 240-time steps. The confusion matrices are computed in a class-wise fashion. The multi-class data are treated as if they were binarised under a one-versus-rest transformation. The y-axis includes the actual class labels and the x-axis shows the algorithm’s predictions. Notice that the different colours represent the percentage range, with white being 100\% and black 0\%.

Figure \ref{fig:confusion_matrix_toilet} presents the confusion matrix of the class label Toilet versus the rest of the classes. The model has difficulty identifying this class as it manages to correctly identify when the toilet was used approximately half of the time. On the other hand, it manages to identify almost all the other events correctly, while misclassifying as “Toilet” only 573 samples.

In Figure \ref{fig:confusion_matrix_shower} we examine the predictions for the class ``Shower''. The model identifies correctly almost 75\% of the samples that are labelled as a shower. Also, it has very few false positives as it distinguishes the rest of the samples with ease.
Figure \ref{fig:confusion_matrix_faucet}, depicts the performance for the class Faucet. We observe that the classifier learnt to identify cases where the faucet was in operation with high accuracy, as it was classified correctly approximately 90\% of the time.

In Figure \ref{fig:confusion_matrix_dish_washer}, we examine the performance of the classifier when identifying the cases where the dish washer was active. As it is evident, the model struggles to classify correctly any of the dish washer samples. Specifically, in 235 (out of the 249) cases only the dishwasher is operating, from which 211 cases are predicted as faucet. In 14 (out of the 249) cases the dishwasher is operating simultaneously with another appliance. In 9 cases (out of the 14) the dishwasher and shower are operating, and the model correctly predicts only the shower. This is attributed partly to the fact that the dishwasher constitutes the minority class with just 1.9\% relative to the other class as shown in Fig. \ref{fig:dataset_appliance_counts}, and partly to the fact that the dish washer cycle exhibits intermittent behavior, thus making it harder for the model to distinguish between the dishwasher and faucet.
    
The results from the predictions for the clothes washer appliance are presented in Figure \ref{fig:confusion_matrix_clothes_washer}. We can observe that the classifier correctly identifies cases where the clothes washer was in operation. Also, we can see that there are very few cases that were misclassified as clothes washers by the model.

Overall, the problem of identifying active appliances from the aggregated water consumption is challenging. We conclude that there are cases which the model performs well, while in others it fails to give accurate predictions. This inconsistency in the performance for some classes can be attributed to three key reasons:
(1) Class imbalance constitutes a key challenge. Recall that the majority of the samples constitute cases where none of the appliances was active (Fig.~\ref{fig:dataset_appliance_counts}). Moreover, the imbalance among the appliances has a significant role, especially for the underrepresented appliances, such as the dish washer. 
(2) Another key factor lies in the aggregated sequence, where some consumption profiles might look similar, especially in the case of appliances, e.g., the dishwasher which has a long and intermittent cycle. The use of a sliding window and the absence of an event detection algorithm to specifically search for these intermittent events, misleads the classifier during inference. 
(3) The simultaneous use of multiple appliances could also yield a water consumption profile which is similar to another.

\section{Conclusions and Future Work}\label{sec:conclusions}
In this work, we have considered an aggregated sequence of a multi-variate time series, and propose a methodology to make predictions based solely on this aggregated information. As a case study, we have considered the challenging problem of non-intrusive water monitoring. The proposed methodology has been demonstrated to be very effective. Identified difficulties are the class imbalance, and the noisy information as a result of the time series aggregation. Future work will attempt to better capture longer temporal correlations using deep neural models, such as, LSTMs and convolutional neural networks \cite{fawaz2019deep}.


%
%
%
\bibliographystyle{splncs04}
\bibliography{mybib}

\begin{thebibliography}{10}
\providecommand{\url}[1]{\texttt{#1}}
\providecommand{\urlprefix}{URL }
\providecommand{\doi}[1]{https://doi.org/#1}

\bibitem{bishop2006}
Bishop, C.M.: Pattern Recognition and Machine Learning (Information Science and
  Statistics). Springer-Verlag, Berlin, Heidelberg (2006)

\bibitem{caruana1997multitask}
Caruana, R.: Multitask learning. Machine Learning  \textbf{28}(1),  41--75
  (1997)

\bibitem{Charalampous2021}
Charalampous, A., Papadopoulos, A., Hadjiyiannis, S., Philimis, P.: {Towards
  hydro-informatics modernization with real-time water consumption
  classification}. IEEE AFRICON Conference  (2021)

\bibitem{chen2016xgboost}
Chen, T., Guestrin, C.: Xgboost: A scalable tree boosting system. In:
  Proceedings of the 22nd ACM SIGKDD International Conference on Knowledge
  Discovery and Data Mining. pp. 785--794 (2016)

\bibitem{cominola2018implications}
Cominola, A., Giuliani, M., Castelletti, A., Rosenberg, D.E., Abdallah, A.M.:
  Implications of data sampling resolution on water use simulation, end-use
  disaggregation, and demand management. Env. Modelling \& Software
  \textbf{102},  199--212 (2018)

\bibitem{cominola2015benefits}
Cominola, A., Giuliani, M., Piga, D., Castelletti, A., Rizzoli, A.E.: Benefits
  and challenges of using smart meters for advancing residential water demand
  modeling and management: A review. Env. Modelling \& Software  \textbf{72},
  198--214 (2015)

\bibitem{cominola2021long}
Cominola, A., Giuliani, M., Castelletti, A., Fraternali, P., Gonzalez, S.L.H.,
  Herrero, J.C.G., Novak, J., Rizzoli, A.E.: Long-term water conservation is
  fostered by smart meter-based feedback and digital user engagement. npj Clean
  Water  \textbf{4}(1),  1--10 (2021)

\bibitem{deoreo2011analysis}
DeOreo, W.B.: Analysis of water use in new single family homes. By Aquacraft.
  For Salt Lake City Corporation and US EPA  (2011)

\bibitem{fawaz2019deep}
Fawaz, H.I., Forestier, G., Weber, J., Idoumghar, L., Muller, P.: Deep learning
  for time series classification: a review. Data Mining and Knowledge Discovery
   \textbf{33}(4),  917--963 (2019)

\bibitem{froehlich2011longitudinal}
Froehlich, J., Larson, E., Saba, E., Campbell, T., Atlas, L., Fogarty, J.,
  Patel, S.: A longitudinal study of pressure sensing to infer real-world water
  usage events in the home. In: International conference on pervasive
  computing. pp. 50--69. Springer (2011)

\bibitem{he2009learning}
He, H., Garcia, E.A.: Learning from imbalanced data. IEEE Transactions on
  Knowledge and Data Engineering  \textbf{21}(9),  1263--1284 (2009)

\bibitem{luaces2012binary}
Luaces, O., D{\'\i}ez, J., Barranquero, J., del Coz, J., Bahamonde, A.: Binary
  relevance efficacy for multilabel classification. Progress in AI
  \textbf{1}(4),  303--313 (2012)

\bibitem{Mazzoni2021}
Mazzoni, F., Alvisi, S., Franchini, M., Ferraris, M., Kapelan, Z.: {Automated
  Household Water End-Use Disaggregation through Rule-Based Methodology}.
  Journal of Water Resources Planning and Management  \textbf{147}(6),
  04021024 (2021)

\bibitem{Nguyen2013}
Nguyen, K.A., Stewart, R.A., Zhang, H.: {An intelligent pattern recognition
  model to automate the categorisation of residential water end-use events}.
  Env. Modelling \& Software  \textbf{47},  108--127 (2013)

\bibitem{ojeda2008classification}
Ojeda~Maga{\~n}a, B., Andina de~la Fuente, D., Nakamura, C., Ruelas, R.:
  Classification of domestic water consumption using an anfis model  (2008)

\bibitem{read2011classifier}
Read, J., Pfahringer, B., Holmes, G., Frank, E.: Classifier chains for
  multi-label classification. Machine Learning  \textbf{85}(3),  333--359
  (2011)

\end{thebibliography}

\end{document}